\documentclass{article}

\usepackage{microtype}
\usepackage{graphicx}
\usepackage{subfigure}
\usepackage{booktabs} % for professional tables

\usepackage{threeparttable} %给表格加注解
\usepackage{bbding}     %实验勾叉
\usepackage{amsmath} %公式对齐
\usepackage{amssymb}

\usepackage{hyperref}

\usepackage[accepted]{icml2021_arxiv}

\icmltitlerunning{Dynamic Normalization}

\begin{document}

\twocolumn[
\icmltitle{Dynamic Normalization}




\icmlsetsymbol{equal}{*}

\begin{icmlauthorlist}
\icmlauthor{Chuan Liu}{to}
\icmlauthor{Yi Gao}{to}
\icmlauthor{Jiancheng Lv}{to}
% \icmlauthor{Iaesut Saoeu}{ed}
% \icmlauthor{Fiuea Rrrr}{to}
% \icmlauthor{Tateu H.~Yasehe}{ed,to,goo}
% \icmlauthor{Aaoeu Iasoh}{goo}
% \icmlauthor{Buiui Eueu}{ed}
% \icmlauthor{Aeuia Zzzz}{ed}
% \icmlauthor{Bieea C.~Yyyy}{to,goo}
% \icmlauthor{Teoau Xxxx}{ed}
% \icmlauthor{Eee Pppp}{ed}
\end{icmlauthorlist}

\icmlaffiliation{to}{Sichuan University, Chengdu 610065, P. R. China}
% \icmlaffiliation{goo}{Googol ShallowMind, New London, Michigan, USA}
% \icmlaffiliation{ed}{School of Computation, University of Edenborrow, Edenborrow, United Kingdom}

\icmlcorrespondingauthor{Chuan Liu}{liu.ca@qq.com}
\icmlcorrespondingauthor{Jiancheng Lv}{lvjiancheng@scu.edu.cn}

\icmlkeywords{Machine Learning, ICML}

\vskip 0.3in
]



\printAffiliationsAndNotice{}  % leave blank if no need to mention equal contribution

\begin{abstract}
Batch Normalization has become one of the essential components in CNN. It allows the network to use a higher learning rate and speed up training. And the network doesn't need to be initialized carefully. However, in our work, we find that a simple extension of BN can increase the performance of the network. First, we extend BN to adaptively generate scale and shift parameters for each mini-batch data, called DN-C (Batch-shared and Channel-wise). We use the statistical characteristics of mini-batch data ($E[X], Std[X]\in\mathbb{R}^{c}$) as the input of SC module. Then we extend BN to adaptively generate scale and shift parameters for each channel of each sample, called DN-B (Batch and Channel-wise). Our experiments show that DN-C model can't train normally, but DN-B model has very good robustness. In classification task, DN-B can improve the accuracy of the MobileNetV2 on ImageNet-100 more than 2\% with only 0.6\% additional Mult-Adds. In detection task, DN-B can improve the accuracy of the SSDLite on MS-COCO nearly 4\% mAP with the same settings. Compared with BN, DN-B has stable performance when using higher learning rate or smaller batch size.
\end{abstract}

\begin{figure}[ht]
\vskip 0.2in
\begin{center}
\centerline{\includegraphics[width=\columnwidth]{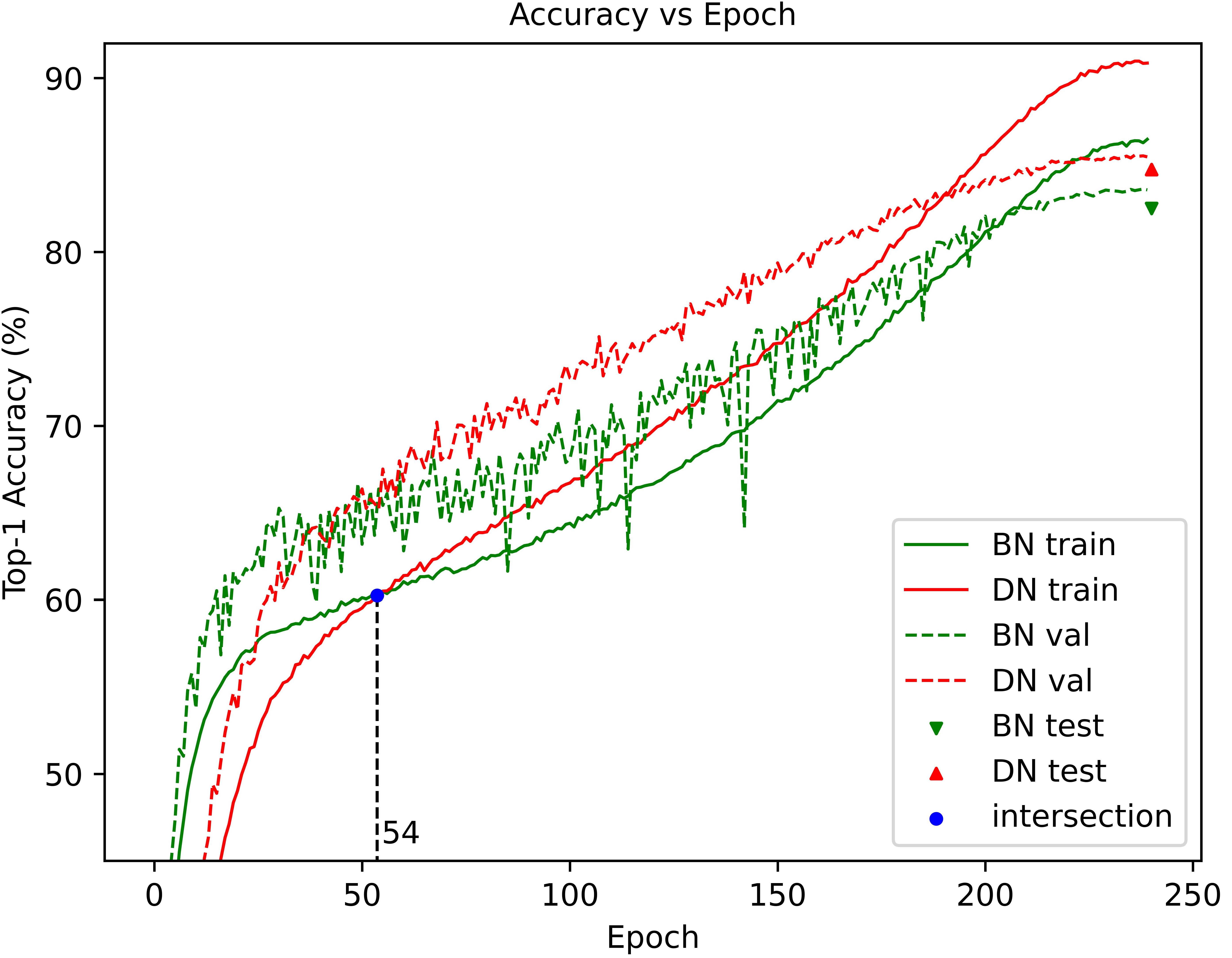}}
\caption{\textbf{Accuracy vs Epoch} on ImageNet-100, using MobileNetV2 (best viewed in color). Compared with BN, DN-B has higher test accuracy and more stable validation accuracy during training. r is 16, g is oup in DN-B.}
\label{fig1}
\end{center}
\vskip -0.2in
\end{figure}

\section{Introduction}
\label{Introduction}
Deep learning has achieved great success in many fields, such as computer vision and natural language processing. In the field of computer vision, there are many famous network architectures. LeNet \cite{lecun1998gradient} network has laid the foundation for the development of CNN (convolutional neural network). AlexNet \cite{krizhevsky2017imagenet} is the winner of the 2012 ImageNet competition. AlexNet uses ReLU \cite{nair2010rectified,jarrett2009best}, Dropout \cite{hinton2012improving,srivastava2014dropout} and other effective tricks, which affects the design of the following network architectures. ResNet \cite{he2016deep} has achieved great success and it proposes an effective strategy to train deeper networks. VGG \cite{simonyan2014very}, GoogleNet \cite{szegedy2015going, ioffe2015batch, szegedy2016rethinking} and other network architectures \cite{huang2017densely, xie2017aggregated} also promote the development of CNN. With the rapid development of CNN model and mobile demand, there are some very powerful lightweight models. MobileNetV1 \cite{howard2017mobilenets} achieves a good balance between accuracy and latency due to depthwise separable convolutions. MobileNetV2 \cite{sandler2018mobilenetv2} utilizes the inverted residual with linear bottleneck. MobileNetV3 \cite{howard2019searching} combined with NAS (network architecture search) has achieved very good results. ShuffleNetV1 \cite{zhang2018shufflenet} uses channel shuffle and group convolution to design the network architecture. ShuffleNetV2 \cite{ma2018shufflenet} provides some guidelines for designing lightweight network structures.

The success of deep learning in computer vision is closely related to the popular components, such as normalization \cite{ioffe2015batch, ba2016layer, wu2018group, ulyanov2016instance}. \cite{ioffe2015batch} find the phenomenon of internal covariate shif and proposes the strategy of normalizing the data of mini-batch. BN \cite{ioffe2015batch} makes it easier to train neural networks, such as using a larger learning rate and it reduces the network initialization requirement. However, BN relies on the statistical characteristics of mini-batch data, the performance of network is not good when the batch size is small. GN (Group Normalization) \cite{wu2018group} keeps its normalization separate from the batch dimension by grouping channels. With such a simple strategy, GN can keep the network performance even when the batch size is small. But the learnable weight parameters in BN and GN are the dimension of channel. In other words, the data in mini-batch share scale and shift parameters. The parameters are static and don't change with the input when the network finishes training.

In addition, some work \cite{hu2018squeeze, yang2019condconv, ma2020weightnet, zhao2020deep, chen2020dynamic} shows that the performance of the dynamic module  is better than the static module in network. SENet \cite{hu2018squeeze} utilizes SE (Squeeze-and-Excitation block) to adaptively learn and calibrate the relationship between channels. SENet has greatly improved the network performance at a small cost and the application scope is very wide. CondConv \cite{yang2019condconv} extends the boundary of convolution kernel, so that convolution kernel can adaptively learn the corresponding convolution kernel weight for each sample. This strategy can improve the performance of the model under the condition of relatively balanced performance and cost. WeightNet \cite{ma2020weightnet} further unifies SENet and CondConv, reducing costs and improving the performance of the model. Dynamic ReLU \cite{chen2020dynamic} and APReLU \cite{zhao2020deep} believe that unified ReLU may limit the learning ability of the model. So Dynamic ReLU and APReLU adaptively learn the corresponding activation function for the sample, which greatly improves the performance of the model.

In this paper, We first let SC-Module adaptively learn the scale and shift parameters of BN \cite{ioffe2015batch}. Then, we extend SC-Module to learn the scale and shift parameters for each sample. Because the scale coefficient in BN can also be considered as a measure of the importance of the channel, we think that Dynamic Normalization also has the channel recalibration ability of SENet \cite{hu2018squeeze} to some extent. And compared with using SENet and BN together, Dynamic Normalization can reduce the number of parameters and maintain the accuracy of the model. In addition, we set two adjustable coefficients, coefficient r  (reduction ratio of channel) and coefficient g (number of cannels per group), to balance the accuracy and cost of the model. Our experiments show that DN-B model has very good robustness.  And DN-B can improve the accuracy of the MobileNetV2 \cite{sandler2018mobilenetv2} on ImageNet-100 \cite{deng2009imagenet} more than 2\% with only 0.6\% additional Mult-Adds. Compared with BN, DN-B has stable performance when using higher learning rate or smaller batch size.

\section{Background and Related Work}
\label{Background and Related Work}
\textbf{Dynamic Network.} Dynamic network has achieved good performance at present. The success of the dynamic network mainly comes from the adaptive adjustment of some components in the network. SENet \cite{hu2018squeeze} improves the performance of the model by adaptively recalibrating the channel for each sample. CondConv \cite{yang2019condconv} enables the convolution kernel weights to be generated adaptively according to each sample. WeightNet \cite{ma2020weightnet} has unified SENet and CondConv in the weight space, reducing the number of parameter and improving the performance of the model. APReLU \cite{zhao2020deep} improves the learning ability of the model and the performance of the model by adaptively learning the activation function of the sample. Dynamic ReLU \cite{chen2020dynamic} proposes three different ways of dynamically learning activation functions: DY-ReLU-A (Spatial and Channel-shared), DY-ReLU-B (Spatial-shared and Channel-wise), DY-ReLU-C (Spatial and Channel-wise).

\begin{figure*}
\vskip 0.2in
\begin{center}
\centerline{\includegraphics[width=0.90\textwidth]{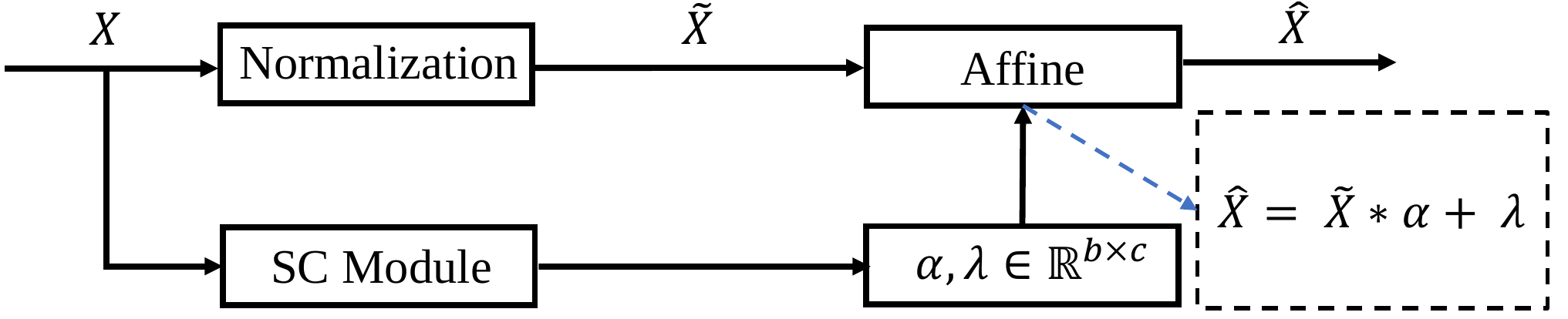}}
\caption{\textbf{The DN-B structure.} The scale and shift parameters are generated by SC-Module. See equation \ref{eq:6} for the details of Normalization module.}
\label{fig2}
\end{center}
\vskip -0.2in
\end{figure*}

\textbf{Attention.} Attention has been widely used in CNN model, such as SENet \cite{hu2018squeeze} and CBAM \cite{woo2018cbam}in computer vision. SENet is the winner of the 2017 ImageNet competition. SENet utilizes Squeeze-and-Excitation block to generate a weight coefficient for each channel of each sample, and then uses these weight coefficients to recalibrate the channels. CBAM uses attention in both channel and spatial dimensions, which achieves better performance than SENet. CBAM \cite{woo2018cbam} adds max-pooling operation, which makes the output of max-pooling and ave-pooling operations share multi-layer perceptron (MLP) with one hidden layer in Channel Attention Module. Then CBAM concats the output of max-pooling and ave-pooling operations, and obtains the spatial attention coefficients through a convolution layer and activation function. And CBAM first recalibrates the channel dimension with Channel Attention Module, and then recalibrates the spatial dimension with Spatial Attention Module.

\textbf{Normalization.} Normalization \cite{ioffe2015batch, ba2016layer, wu2018group, ulyanov2016instance} is one of the essential components in deep learning, which promotes the development of deep learning. As one of the most commonly used normalization techniques in computer vision, BN \cite{ioffe2015batch} can speed up the training of neural network and avoid careful initialization to some extent. However, BN uses the statistical characteristics of mini-batch data. When the batch size is small, the network performance using BN is not good. So GN \cite{wu2018group} groups channels so that it is independent on the batch dimension. However, the scale and shift parameters of BN and GN are both channel dimensions and are shared in batch dimension. In other words, BN and GN are batch-shared and channel-wise. In this work, We first extend normalization to generate parameters adaptively for mini-batch data (DN-C, Batch-shared and Channel-wise). Then, we extend DN-B to generate parameters adaptively for each channel of each sample (DN-B, Batch and Channel-wise).

\section{Dynamic Normalization}
\label{Dynamic Normalization}
\subsection{Batch Normalization} In this part, We will give a brief introduction to BN.
% \textbf{Normalizattion.}

\begin{equation}
X = \frac{x-E[x]}{\sqrt{Var[x] + \epsilon}}
\end{equation}

where x denotes the input features. $E[x]$ denotes the mean of $x$, and $Var[x]$ denotes the variance of $x$, $E[x]$, $Var[x]$ $\in$ $\mathbb{R}^{c}$. $\epsilon$ is the minimum to prevent division by zero.

\textbf{Affine.}

\begin{equation}
y = X * \gamma + \beta
\end{equation}
where $\gamma$ and $\beta$ are learnable parametes, $\gamma,\beta$ $\in$ $\mathbb{R}^{c}$, and c denotes the channel number of the input feature.

\subsection{SENet and WeightNet}
Squeeze-and-Excitation (SE) block \cite{hu2018squeeze} is powerful and plug and play. SE first uses pooling operation to compress the input features in spatial dimension. And uses a fully connected layer to compress the features in channel dimension. Then a fully connected layer is used to restore the channel dimension. Finally, the final channel coefficients are obtained through Sigmoid function, and the channels are recalibrated with the coefficients.

WeightNet \cite{ma2020weightnet} improves the performance of the network by unifying SE \cite{hu2018squeeze} and CondConv \cite{yang2019condconv} in the weight space. WeightNet replaces the two fully connected layers in SE with 1 * 1 convolution operations (FC and Grouped FC). By using convolution, the grouped fully connected operation can be realized. At the same time, it also reduces the amount of parameters. Through these strategies, WeightNet can adaptively generate weight parameters for each sample. It enhances the learning ability of the network, and can well balance the cost and accuracy of the model. In our implementation of SC-Module, we also use FC and Grouped FC to generate our scale and shift parameters.
 
\subsection{Dynamic Normalization Structure}

In this section, we will  introduce the Dynamic Normalization with batch dimension (DN-B), without batch dimension (DN-C). If we use SE block before BN, we can get equation \ref{eq:3}, \ref{eq:5} and 5.
\begin{equation}\label{eq:3}
x = \alpha * X
\end{equation}
% \begin{equation}\label{eq:5}
% y = \frac{x-E[\alpha * X]}{\sqrt{Var[\alpha * X] + \epsilon}} * \gamma + \beta
% \end{equation}
\begin{align}\label{eq:5}
y &= \frac{x-E[x]}{\sqrt{Var[x] + \epsilon}} * \gamma + \beta \\
  & = \frac{x-E[\alpha * X]}{\sqrt{Var[\alpha * X] + \epsilon}} * \gamma + \beta
\end{align}

where $X$ is the output feature through the convolution kernels, $\alpha$ denotes the the importance of each channel derived from the SE-Module.  $\gamma,\beta$ $\in$ $\mathbb{R}^{c}$, and c denotes the channel number of the input feature.

\begin{equation}\label{eq:6}
\widetilde{X} = \frac{X-E[X]}{\sqrt{Var[X] + \epsilon}}
\end{equation}

\begin{equation}\label{eq:7}
\widehat{X} = \widetilde{X} * \alpha + \lambda
\end{equation}
where $X$ is the output feature through the convolution kernels, $\alpha$ denotes the the scale parameter of each channel derived from the SC-Module. In DN-B, $\alpha,\lambda$ $\in$ $\mathbb{R}^{n,c}$, n denotes the batch size, and c denotes the channel number of the input feature. And in DN-C, $\alpha,\lambda$ $\in$ $\mathbb{R}^{c}$. See Figure \ref{fig2} for details of DN.

\textbf{DN-B and DN-C.}
First, we extend BN to adaptively generate scale and shift parameters for each mini-batch data, called DN-C (Batch-shared and Channel-wise). We use the statistical characteristics of mini-batch data ($E[X],
Std[X]\in\mathbb{R}^{c}$) as the input of SC-Module. We can adaptively generate scale and shift parameters through SC-Module. Finally, through the Affine operation, we get final output.

Then we extend BN to adaptively generate scale and shift parameters for each channel of each sample, called DN-B (Batch and Channel-wise). We use the original input features as the input of SC-Module to generate the scale and shift parameters ($\alpha$, $\lambda$ $\in$ $\mathbb{R}^{n,c}$). Finally, through the Affine operation, we get final output. See Figure \ref{fig2} for details of DN-B.

\section{Experiments}
%\subsection{MobileNetV2}
In this section, we evaluate DN operation on ImageNet-100 dataset\protect\footnote{We randomly select 100 classes in the ImageNet dataset~\cite{deng2009imagenet} called ImageNet-100, where twenty percent of the selected data is used as the validation set. And the original validation set is used as the test set.}~\cite{deng2009imagenet} and MS-COCO dataset~\cite{lin2014microsoft}. On the ImageNet-100 dataset, we chose the lightweight MobileNetV2~\cite{sandler2018mobilenetv2} and ResNet-18/34 as the baseline network. On the MS-COCO dataset, we chose the lightweight detection system, SSDLite~\cite{liu2016ssd}. And we use MobileNetV2 as the backbone network of SSDLite. 

\begin{table}[ht]
\caption{The accuracy of the MobileNetV2 with different modules on ImageNet-100.}\label{tab1}
\vskip 0.15in
\centering
\begin{threeparttable}
\setlength{\tabcolsep}{0.6mm}{
\begin{tabular}{c|c|c|c|c|c|c|c}
\hline
\multicolumn{8}{c}{batch size=64, epoch=240, Lr=0.2} \\
\hline
Module & r & g & \#p & Mult-Adds & Train & Val & Test\\
\hline
BN & \XSolid & \XSolid & 2.35M & 299.62M & 86.48 & 83.61 & 82.50 \\
DN-C\tnote{1} & 16 & \XSolid  & 3.03M & 301.02M & NAN& NAN & \XSolid\\
DN-C\tnote{2} & 16 & \XSolid  & 3.03M & 301.05M & NAN & NAN & \XSolid\\
DN-B & 16 & 1  & 3.03M & 300.34M & 91.62 & 85.63 & 84.08 \\
DN-B\tnote{3} & 16 & oup & 4.36M & 301.66M & 90.86 & 85.57 & 84.74 \\
\hline
\end{tabular}}
\begin{tablenotes}
       \footnotesize
       \item[1] Add the $E[X]$ branch  and the $Std[X]$ branch after the first FC. The learning rate is 0.1.  
       \item[2] Add the $E[X]$ branch  and the $Std[X]$ branch after the second FC. The learning rate is 0.1. 
       \item[3] The SC-Module in this experiment is implemented by two full connection operations.
     \end{tablenotes}
    \end{threeparttable}
\vskip -0.1in
\end{table}

\begin{table}[h]
\caption{The accuracy of the MobileNetV2 with different modules on ImageNet-100.}\label{tab2}
\vskip 0.15in
\centering
\begin{threeparttable}
\setlength{\tabcolsep}{0.5mm}{
\begin{tabular}{c|c|c|c|c|c|c|c|c}
\hline
\multicolumn{9}{c}{batch size=64, epoch=240, Lr=0.2} \\
\hline
Module & r & g & Lr & \#p & Mult-Adds & Train& Val& Test\\
\hline
BN & \XSolid & \XSolid & 0.2 & 2.35M & 299.62M & 86.48 & 83.61 & 82.50\\
SE & 16 & \XSolid & 0.2 & 3.73M & 300.98M & 89.96 & 83.90 & 82.00\\
DN-B & 16 & 1 &   0.2 & 3.03M & 300.34M & 91.62& 85.63 & 84.08 \\
DN-B\tnote{1}& 16 & oup & 0.2 & 4.36M & 301.66M & 90.86 & 85.57 & 84.74\\
\hline
\end{tabular}}
\begin{tablenotes}
       \footnotesize
       \item[1] The SC-Module in this experiment is implemented by two full connection operations.
     \end{tablenotes}
    \end{threeparttable}
\vskip -0.1in
\end{table}

\begin{table}
\caption{The accuracy of the ResNet with different modules on ImageNet-100.}\label{tab3}
\vskip 0.15in
\centering
\begin{threeparttable}
\setlength{\tabcolsep}{0.6mm}{
\begin{tabular}{c|c|c|c|c|c|c|c|c}
\hline
\multicolumn{9}{c}{ResNet18/ResNet-34, Lr=learning rate, b=batch size} \\
\hline
res-18 & r & g & Lr & \#p & Mult-Adds & Train& Val& Test\\
\hline
BN & \XSolid & \XSolid & 0.2 & 11.22M & 1730.12M & 94.57 & 85.89 & 85.00\\
BN & \XSolid & \XSolid & 0.5 & 11.22M & 1730.12M & 93.35 & 86.07 & 84.64\\
BN & \XSolid & \XSolid & 0.8 & 11.22M & 1730.12M & 88.61 & 84.75 & 82.88\\
% BN & \XSolid & \XSolid & 8 & 11.22M & 1730.12M &  &  & \\
DN-B & 4 & oup & 0.2 & 12.52M & 1731.43M & 94.19 & 85.58 & 84.36\\
DN-B & 4 & oup & 0.5 & 12.52M & 1731.43M & 93.05 & 85.75 & 85.34\\
DN-B & 16 & oup & 0.8 & 11.54M  & 1730.45M & 92.56 & 85.85 & 84.78 \\
% DN-B & 16 & oup & 8 &  &  &  &  & \\
\hline
res-34 & r & g & b & \#p & Mult-Adds & Train& Val& Test\\
\hline
\tnote{2} BN & \XSolid & \XSolid & 64 & 21.33M & 3579.81M & 94.98 & 86.89 & 85.84\\
\tnote{3} BN & \XSolid & \XSolid & 8 & 21.33M & 3579.81M & 83.76 & 83.51 & 82.32\\
\tnote{4} BN & \XSolid & \XSolid & 4 & 21.33M & 3579.81M & 47.79 & 54.10 & 53.20 \\
\tnote{4} DN-B & 16 & oup & 4 & 21.85M & 3580.35M & 75.14 & 78.41 & 77.10\\
\tnote{3} DN-B & 16 & oup & 8 & 21.85M & 3580.35M & 84.86 & 84.65 & 83.50\\
\tnote{2} DN-B & 16 & oup & 64 & 21.85M & 3580.35M & 95.02 & 86.97 & 85.42\\
\tnote{2} DN-B & 32 & oup & 64 & 21.58M & 3580.08M & 95.17 & 86.72 & 85.96 \\
\hline
\end{tabular}}
\begin{tablenotes}
       \footnotesize
       \item[1] The SC-Module in this experiment is implemented by two full connection operations.
       \item[2] The models take 240 epochs.
       \item[3] The models only take 120 epochs.
       \item[4] The models only take 60 epochs.
     \end{tablenotes}
    \end{threeparttable}
\vskip -0.15in
\end{table}

\subsection{The performance of MobileNetV2 on ImageNet-100}
Our implementation code is based on DenseNAS~\cite{fang2020densely}. In order to prove the effectiveness of our method, we directly replace BN operation in MobileNetV2 with DN operation. First, we want BN to be dynamic, so we use the statistical characteristics of data in mini-batch as the input of SC-Module. But from table~\ref{tab1}, When we use DN-C directly, we find that the model can't train normally. Therefore, we further extend DN-C to generate scale and shift parameters for each sample. From table~\ref{tab2}, we can see that DN-B improves the accuracy of the model with less parameter cost. At the same time, we use SE block directly in front of BN, which reduces the accuracy of the model, thus proving that our method is not just an extension of SE block. In addition, we also study the influence of other parameters on MobileNetV2 with DN-B, such as r, g, learning rate and batch size. We find that compared with BN, DN-B has better robustness, and the model can still maintain better accuracy even when using the higher learning rate or smaller batch size. For details of the experiment, see ~\ref{Ablation study and Analysis}. 

\begin{table*}[ht]
\caption{The influence of batch size.}\label{tab8}
\vskip 0.15in
\centering
\begin{threeparttable}
\setlength{\tabcolsep}{4.5mm}{
\begin{tabular}{c|c|c|c|c|c|c}
\hline
\multicolumn{7}{c}{MobileNetV2, ImageNet-100, Lr=0.2, DN-B} \\
\hline
BN/DN-B & r and g & \#p & Mult-Adds & Train acc(last) & Val acc(best) & Test acc \\
\hline
BN-8\tnote{1} & \XSolid & 2.35M & 299.62M& 1.00/4.95 & 2.50/12.04 & \XSolid \\
DN-8\tnote{1} & 16, 1 &3.03M & 300.34M& 73.81/90.81 & 79.60/94.01 & 78.42/92.92 \\
BN-64\tnote{2} & \XSolid &2.35M & 299.62M & 86.48/96.45 & 83.61/95.54 & 82.50/94.80\\
DN-64\tnote{2} & 16, 1 & 3.03M & 300.34M & 91.62/97.74 & 85.63/95.92 & 84.08/94.78 \\
\hline
\multicolumn{7}{c}{MobileNetV2, ImageNet-100, Lr=0.2, DN-B} \\
\hline
BN/DN-B & r and g & \#p & Mult-Adds & Train acc(last) & Val acc(best) & Test acc \\
\hline
BN-64-1\tnote{3} & \XSolid & 2.35M & 299.62M& 86.48/96.45 & 83.61/95.54 & 82.50/94.80 \\
DN-64-1\tnote{3} & 16, 1 &3.03M & 300.34M& 73.81/90.81 & 79.60/94.01 & 83.92/94.94 \\
BN-64-2 & \XSolid & 2.35M & 299.62M& 86.48/96.45 & 83.61/95.54 & 82.50/94.80 \\
DN-64-2 & 16, 1 &3.03M & 300.34M& 91.62/97.74 & 85.63/95.92 &  84.02/94.90\\
BN-64-32  & \XSolid & 2.35M & 299.62M& 86.48/96.45 & 83.61/95.54 & 82.50/94.80 \\
DN-64-32 & 16, 1 &3.03M & 300.34M& 91.62/97.74 & 85.63/95.92 & 84.08/94.78 \\
\hline
\end{tabular}}
\begin{tablenotes}
       \footnotesize
       \item[1] BN-8 and DN-8 denote the training bacth size is 8. The model takes 120 epochs.
       \item[2] BN-64 and DN-64 denote the training bacth size is 64. The model takes 240 epochs.
       \item[3] BN-64-1 and DN-64-1 denote the training batch size is 64, test batch size is 1.
     \end{tablenotes}
    \end{threeparttable}
\vskip -0.1in
\end{table*}

\subsection{The performance of ResNet on ImageNet-100}
Our implementation code is based on DenseNAS~\cite{fang2020densely}. Using the higher learning rate on ResNet-18, we find that the model with DN can maintain better accuracy. In addition, we use smaller batch size on ResNet-34 and find that the model using DN has higher accuracy. These experiments are consistent with the performance of DN in MobileNetV2, which proves the robustness of DN. 
% However, we also find that the accuracy of the model will be slightly decreased if we directly replace all BN operations in ResNet.

\begin{table}[h]
\caption{Object detection on MS-COCO.}\label{tab4}
\vskip 0.15in
\centering
\begin{threeparttable}
\setlength{\tabcolsep}{1.5mm}{
\begin{tabular}{c|c|c|c|c|c|c}
\hline
\multicolumn{7}{c}{batch size=64, GPU=2} \\
\hline
Module & r & g & epoch & \#p & MAdds & mAP(\%)\\
\hline
BN & \XSolid & \XSolid & 30 & 4.32M & 0.81GB & 15.3\\
DN-B\tnote{1} & 16 & oup & 30 & 6.33M & 0.81GB & 19.2 \\
\hline
\end{tabular}}
\begin{tablenotes}
       \footnotesize
       \item[1] The SC-Module in this experiment is implemented by two full connection operations.
     \end{tablenotes}
    \end{threeparttable}
\vskip -0.1in
\end{table}

\subsection{Object Detection}
\begin{figure}[h]
\vskip 0.06in
\begin{center}
\centerline{\includegraphics[width=\columnwidth]{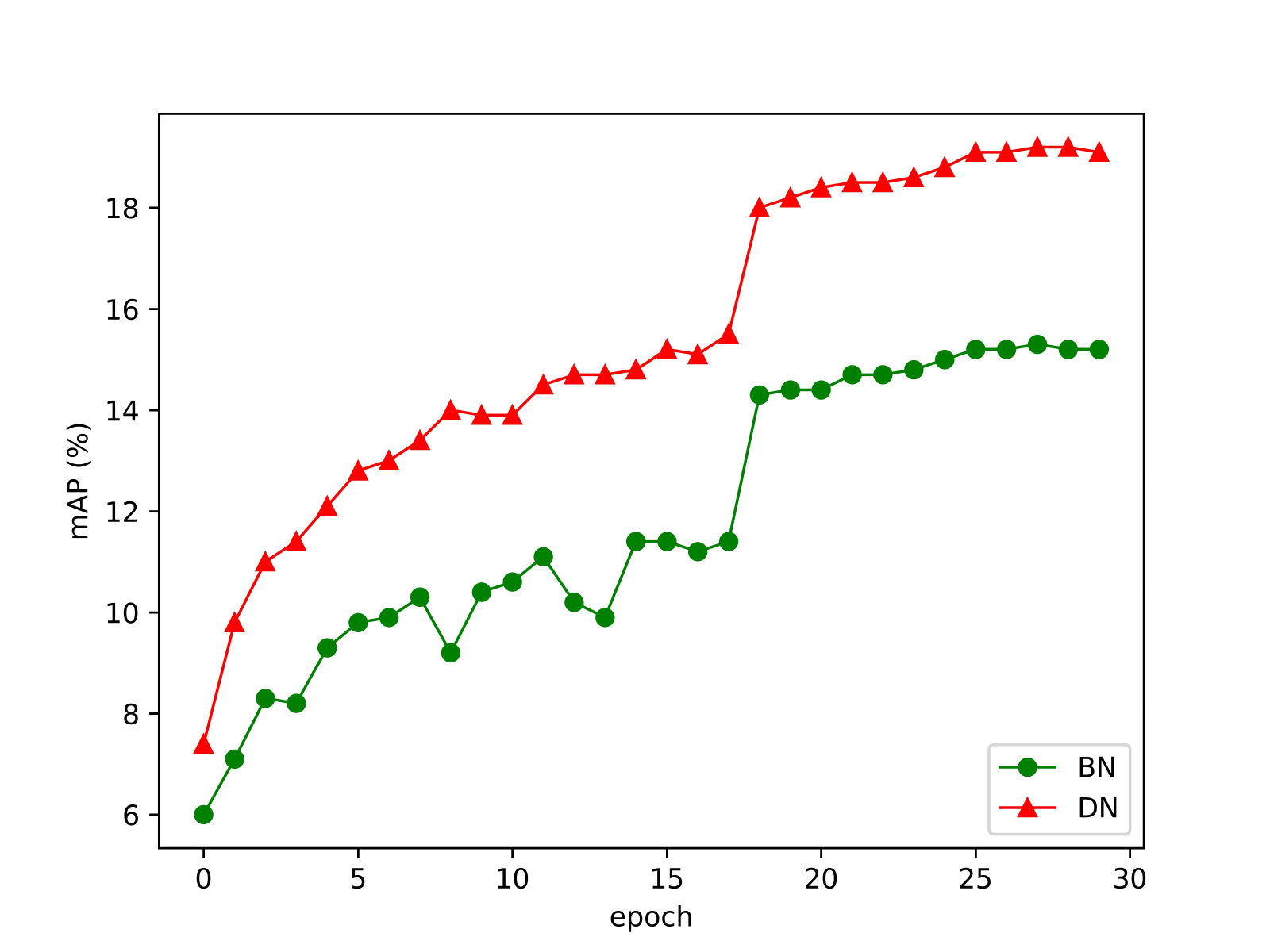}}
\caption{\textbf{mAP vs Epoch} on MS-COCO, using SSDLite (best viewed in color). Compared with BN, DN-B has better accuracy. r is 16, g is oup in DN-B.}
\label{fig3}
\end{center}
\vskip -0.2in
\end{figure}

We verify the effectiveness of our method on the MS-COCO dataset~\cite{lin2014microsoft}. We use a lightweight detection system, SSDLite~\cite{liu2016ssd, sandler2018mobilenetv2}. For convenience, our code is based on MMDetection~\cite{chen2019mmdetection} and FNA~\cite{fang2020fast}. First, we train the MobileNetV2-BN and the MobileNetV2-DN-B with the same settings on ImageNet-100. Then, we use these two models as the backbone network of SSDLite. To make a fair comparison, instead of SyncBN, we use BN. We only train 30 epochs, the initial learning rate is 0.05 and decays at 18, 25, 28 epochs. The batch size of each GPU is 64. From the tabel \ref{tab4} and figure \ref{fig3}, we can see using DN can get better accuracy. Note that all settings are the same except that the backbone network uses DN-B operation instead of BN operation.

\subsection{Ablation study and Analysis}
\label{Ablation study and Analysis}
We set two coefficients in SC-Module to balance accuracy and cost of the module. In this part, we give the results of models when using different values of r and g. In addition, we also give the results of larger learning rate and different batch size.

\begin{figure*}[ht]
\vskip 0.2in
\subfigure{
\begin{minipage}[t]{0.05\linewidth}
% \centering
\includegraphics[width=0.6in, height=0.6in]{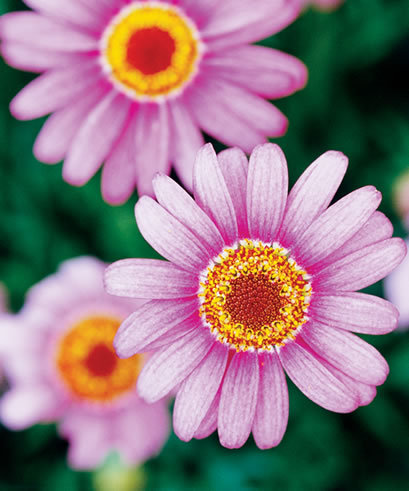}
%\caption{fig2}
\end{minipage}
}%
\hspace{.23in}
\subfigure{
\begin{minipage}[t]{0.05\linewidth}
% \centering
\includegraphics[width=0.6in, height=0.6in]{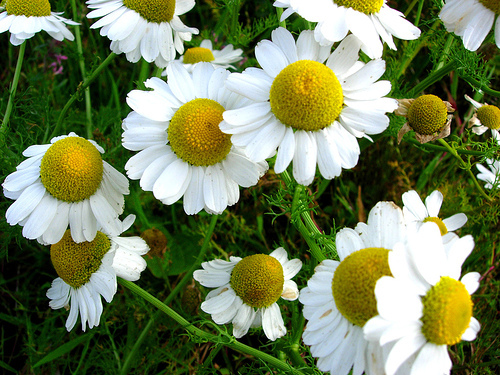}
%\caption{fig2}
\end{minipage}
}%
\hspace{.23in}
% \quad %这个回车键很重要 \quad也可以
\subfigure{
\begin{minipage}[t]{0.05\linewidth}
% \centering
\includegraphics[width=0.6in, height=0.6in]{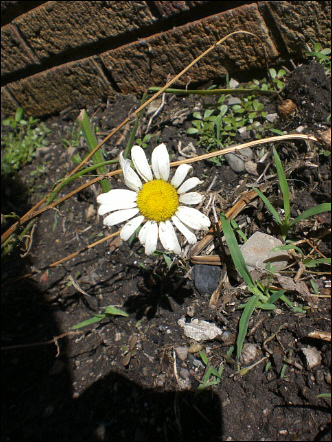}
%\caption{fig2}
\end{minipage}
}%
\hspace{.23in}
\subfigure{
\begin{minipage}[t]{0.05\linewidth}
% \centering
\includegraphics[width=0.6in, height=0.6in]{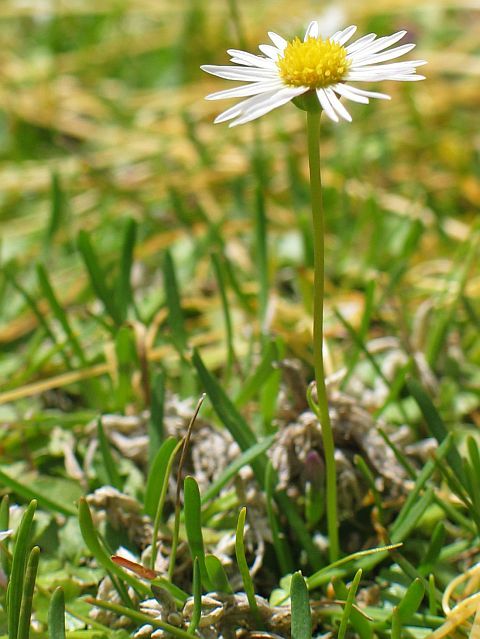}
%\caption{fig2}
\end{minipage}
}%
\hspace{.23in}
\subfigure{
\begin{minipage}[t]{0.05\linewidth}
% \centering
\includegraphics[width=0.6in, height=0.6in]{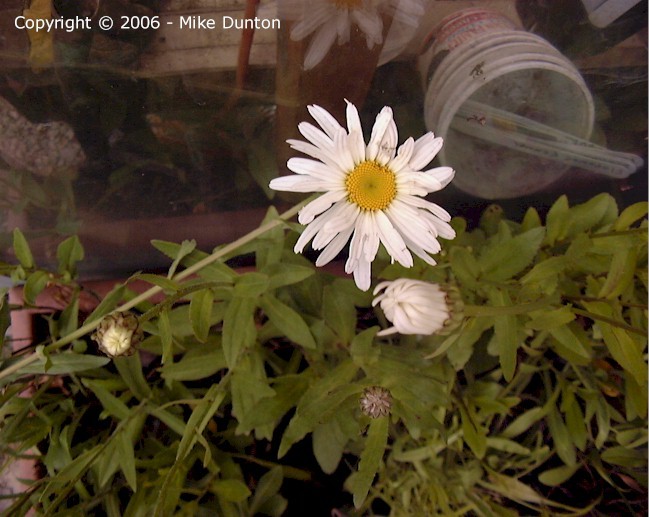}
%\caption{fig2}
\end{minipage}
}%
\hspace{.23in}
\subfigure{
\begin{minipage}[t]{0.05\linewidth}
% \centering
\includegraphics[width=0.6in, height=0.6in]{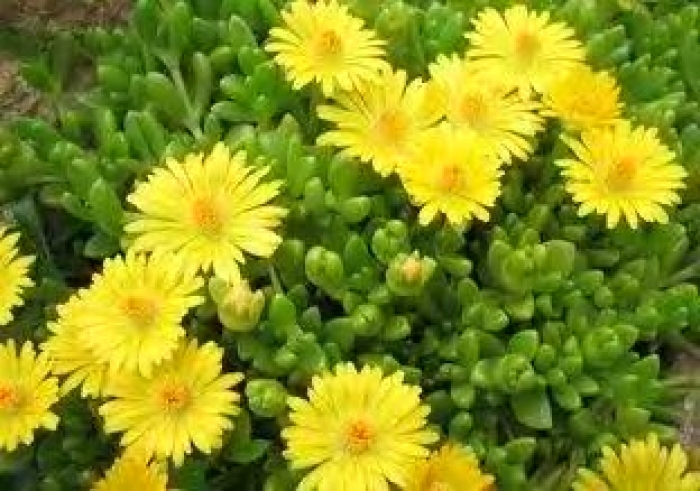}
%\caption{fig2}
\end{minipage}
}%
\hspace{.23in}
\subfigure{
\begin{minipage}[t]{0.05\linewidth}
% \centering
\includegraphics[width=0.6in, height=0.6in]{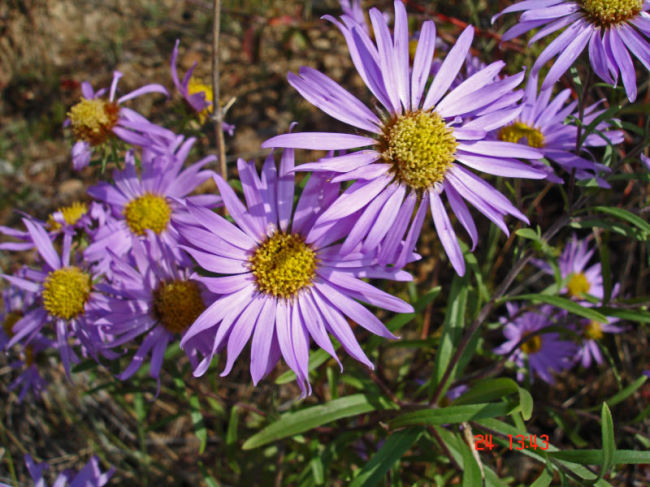}
%\caption{fig2}
\end{minipage}
}%
\hspace{.23in}
\subfigure{
\begin{minipage}[t]{0.05\linewidth}
% \centering
\includegraphics[width=0.6in, height=0.6in]{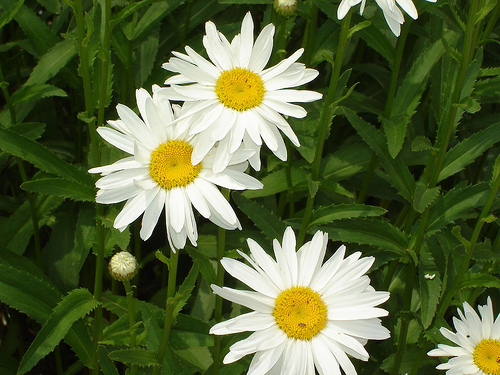}
%\caption{fig2}
\end{minipage}
}%
\hspace{.23in}
\subfigure{
\begin{minipage}[t]{0.05\linewidth}
% \centering
\includegraphics[width=0.6in, height=0.6in]{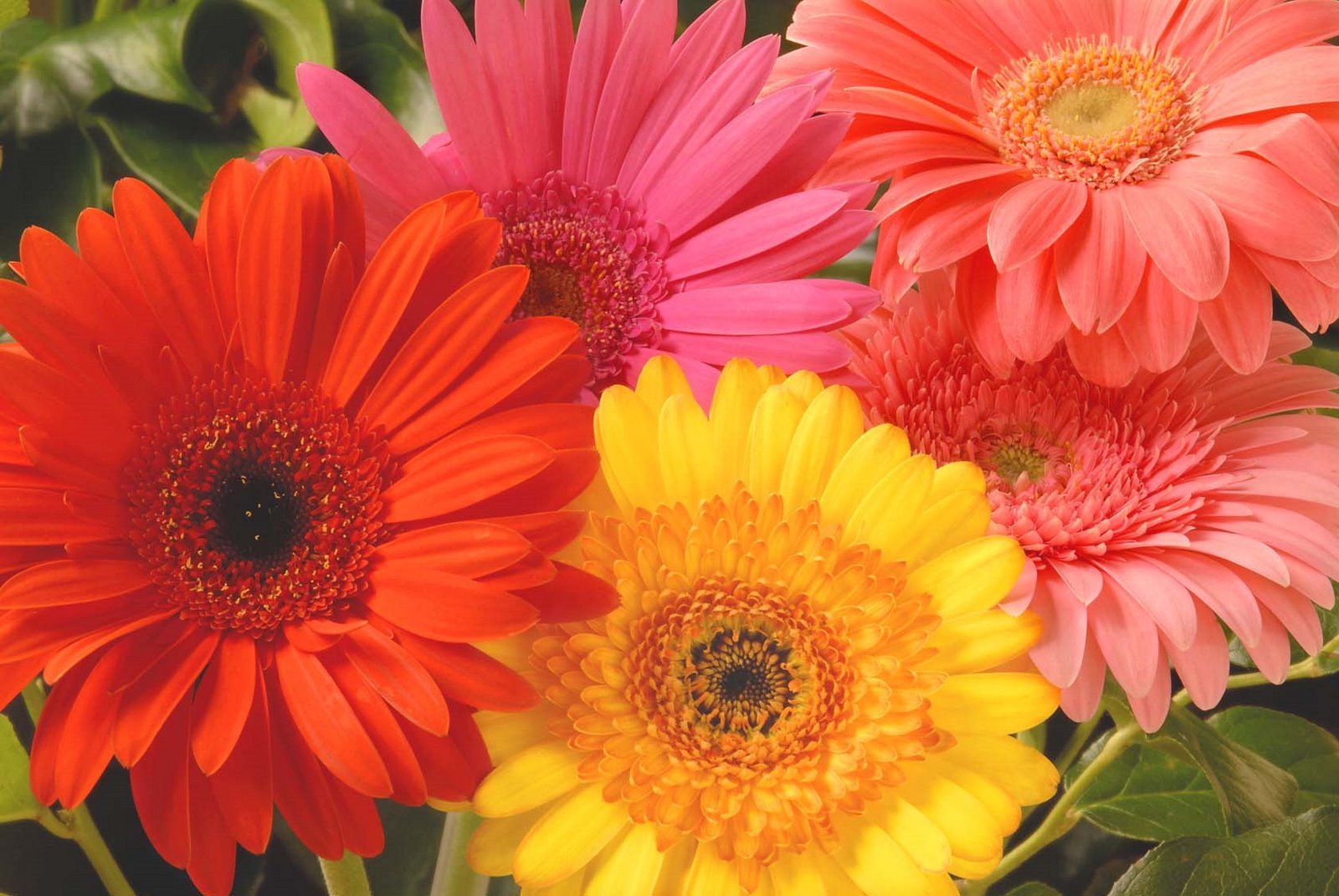}
%\caption{fig2}
\end{minipage}
}%
\hspace{.23in}
\subfigure{
\begin{minipage}[t]{0.05\linewidth}
% \centering
\includegraphics[width=0.6in, height=0.6in]{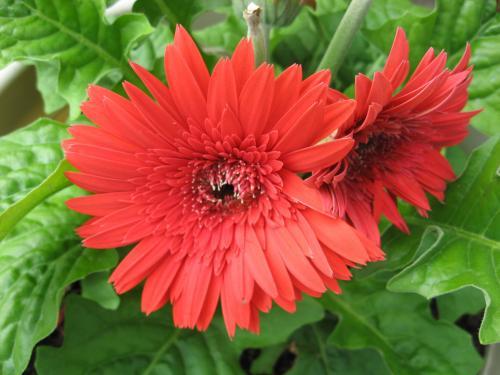}
%\caption{fig2}
\end{minipage}
}%
\\ 
% 2222222222222
\hspace{.3in} 
\subfigure{
\begin{minipage}[t]{0.05\linewidth}
% \centering
\includegraphics[width=0.6in, height=0.6in]{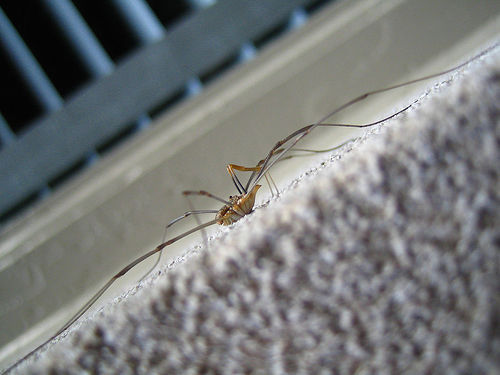}
%\caption{fig2}
\end{minipage}
}%
\hspace{.23in} 
\subfigure{
\begin{minipage}[t]{0.05\linewidth}
% \centering
\includegraphics[width=0.6in, height=0.6in]{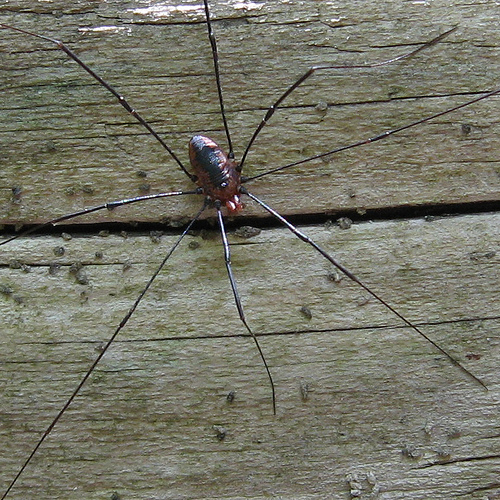}
%\caption{fig2}
\end{minipage}
}%
\hspace{.23in}
% \quad %这个回车键很重要 \quad也可以
\subfigure{
\begin{minipage}[t]{0.05\linewidth}
% \centering
\includegraphics[width=0.6in, height=0.6in]{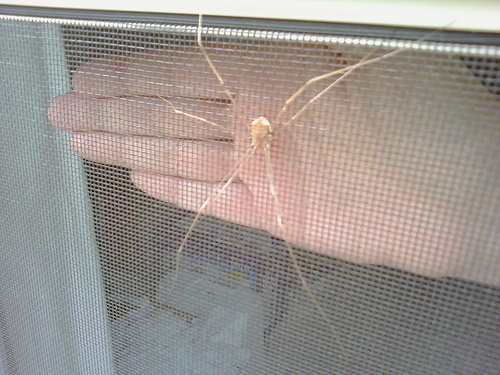}
%\caption{fig2}
\end{minipage}
}%
\hspace{.23in}
\subfigure{
\begin{minipage}[t]{0.05\linewidth}
% \centering
\includegraphics[width=0.6in, height=0.6in]{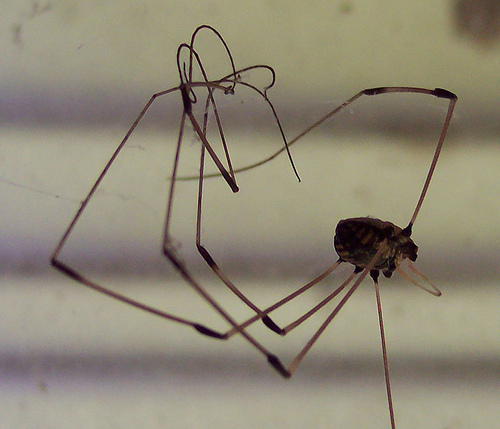}
%\caption{fig2}
\end{minipage}
}%
\hspace{.23in} 
\subfigure{
\begin{minipage}[t]{0.05\linewidth}
% \centering
\includegraphics[width=0.6in, height=0.6in]{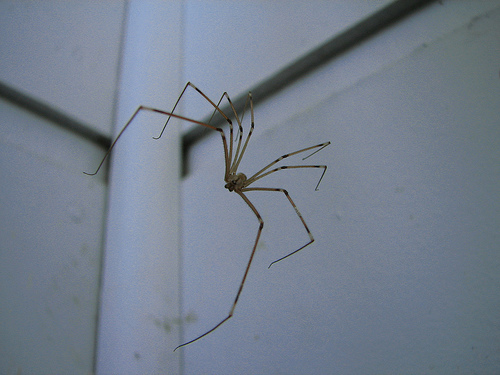}
%\caption{fig2}
\end{minipage}
}%
\hspace{.23in} 
\subfigure{
\begin{minipage}[t]{0.05\linewidth}
% \centering
\includegraphics[width=0.6in, height=0.6in]{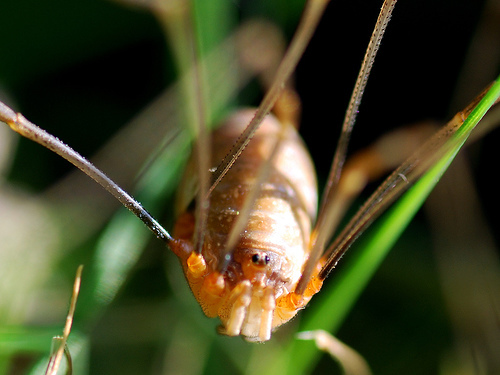}
%\caption{fig2}
\end{minipage}
}%
\hspace{.23in}
\subfigure{
\begin{minipage}[t]{0.05\linewidth}
% \centering
\includegraphics[width=0.6in, height=0.6in]{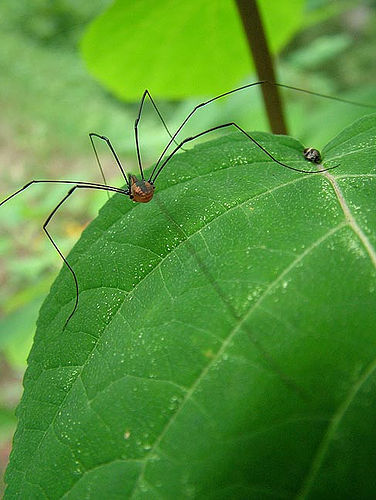}
%\caption{fig2}
\end{minipage}
}%
\hspace{.23in}
\subfigure{
\begin{minipage}[t]{0.05\linewidth}
% \centering
\includegraphics[width=0.6in, height=0.6in]{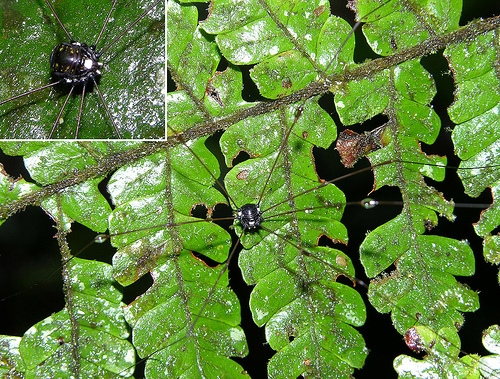}
%\caption{fig2}
\end{minipage}
}%
\hspace{.23in} 
\subfigure{
\begin{minipage}[t]{0.05\linewidth}
% \centering
\includegraphics[width=0.6in, height=0.6in]{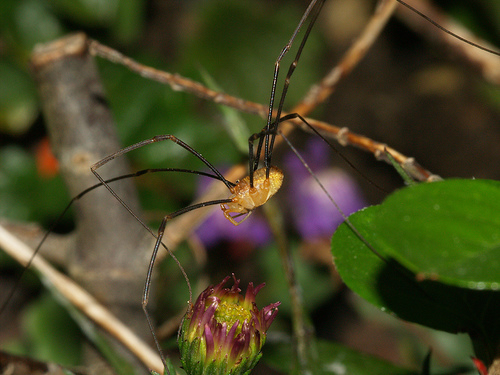}
%\caption{fig2}
\end{minipage}
}%
\hspace{.23in}
\subfigure{
\begin{minipage}[t]{0.05\linewidth}
% \centering
\includegraphics[width=0.6in, height=0.6in]{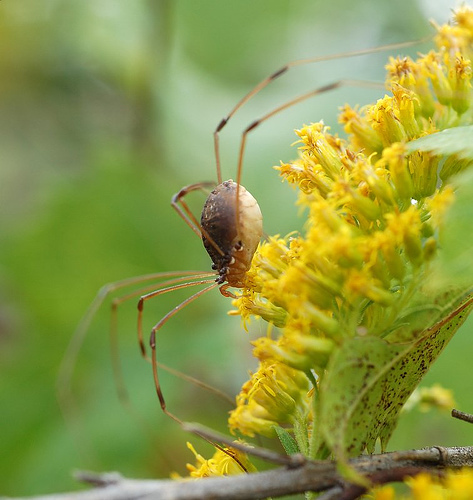}
%\caption{fig2}
\end{minipage}
}%
\\
% 3333333333333333333
\hspace{.3in} 
\subfigure{
\begin{minipage}[t]{0.05\linewidth}
% \centering
\includegraphics[width=0.6in, height=0.6in]{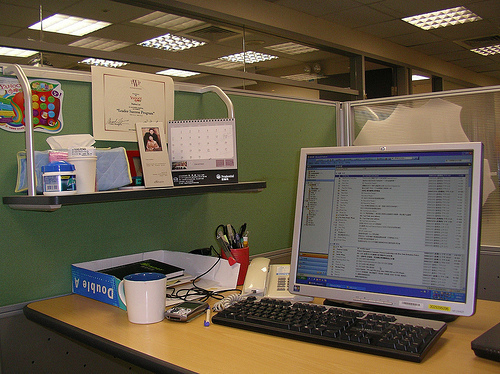}
%\caption{fig2}
\end{minipage}
}%
\hspace{.23in}
\subfigure{
\begin{minipage}[t]{0.05\linewidth}
% \centering
\includegraphics[width=0.6in, height=0.6in]{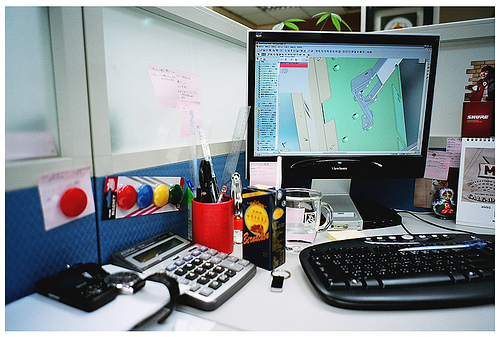}
%\caption{fig2}
\end{minipage}
}%
\hspace{.23in}
% \quad %这个回车键很重要 \quad也可以
\subfigure{
\begin{minipage}[t]{0.05\linewidth}
% \centering
\includegraphics[width=0.6in, height=0.6in]{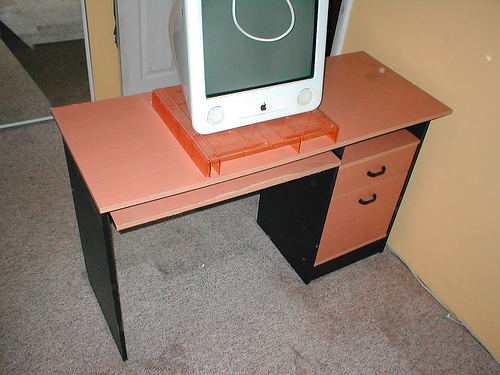}
%\caption{fig2}
\end{minipage}
}%
\hspace{.23in}
\subfigure{
\begin{minipage}[t]{0.05\linewidth}
% \centering
\includegraphics[width=0.6in, height=0.6in]{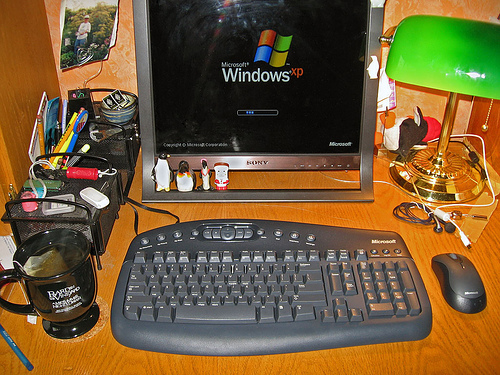}
%\caption{fig2}
\end{minipage}
}%
\hspace{.23in}
\subfigure{
\begin{minipage}[t]{0.05\linewidth}
% \centering
\includegraphics[width=0.6in, height=0.6in]{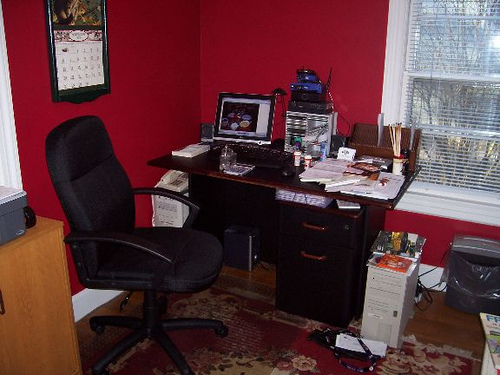}
%\caption{fig2}
\end{minipage}
}%
\hspace{.23in} 
\subfigure{
\begin{minipage}[t]{0.05\linewidth}
% \centering
\includegraphics[width=0.6in, height=0.6in]{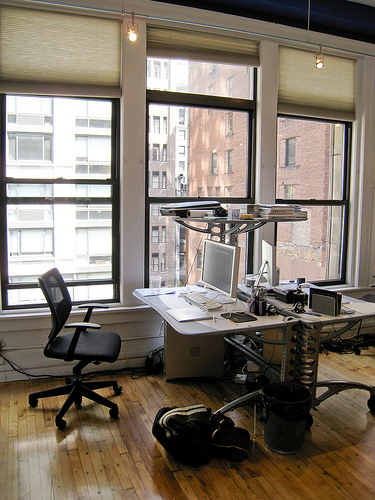}
%\caption{fig2}
\end{minipage}
}%
\hspace{.23in}
\subfigure{
\begin{minipage}[t]{0.05\linewidth}
% \centering
\includegraphics[width=0.6in, height=0.6in]{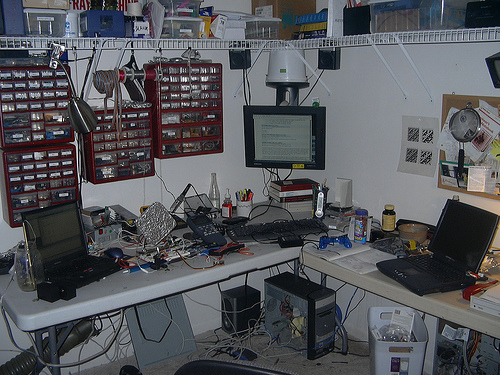}
%\caption{fig2}
\end{minipage}
}%
\hspace{.23in}
\subfigure{
\begin{minipage}[t]{0.05\linewidth}
% \centering
\includegraphics[width=0.6in, height=0.6in]{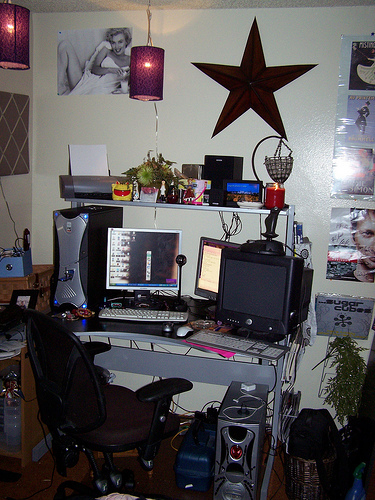}
%\caption{fig2}
\end{minipage}
}%
\hspace{.23in}
\subfigure{
\begin{minipage}[t]{0.05\linewidth}
% \centering
\includegraphics[width=0.6in, height=0.6in]{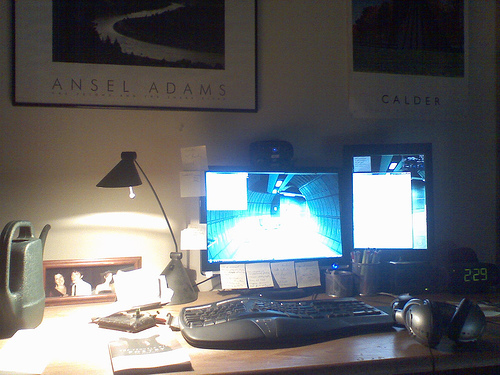}
%\caption{fig2}
\end{minipage}
}%
\hspace{.23in}
\subfigure{
\begin{minipage}[t]{0.05\linewidth}
% \centering
\includegraphics[width=0.6in, height=0.6in]{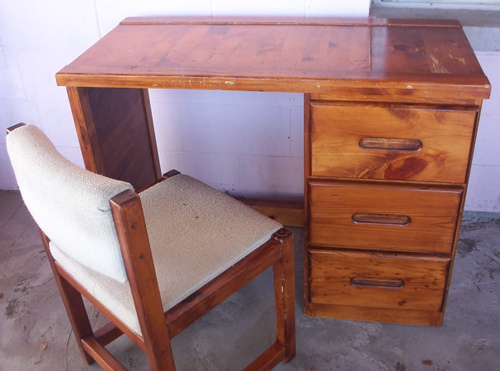}
%\caption{fig2}
\end{minipage}
}% 444444444444444444444444
\\
\hspace{.3in} 
\subfigure{
\begin{minipage}[t]{0.05\linewidth}
% \centering
\includegraphics[width=0.6in, height=0.6in]{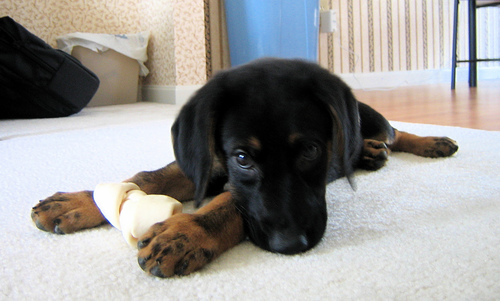}
%\caption{fig2}
\end{minipage}
}%
\hspace{.23in}
\subfigure{
\begin{minipage}[t]{0.05\linewidth}
% \centering
\includegraphics[width=0.6in, height=0.6in]{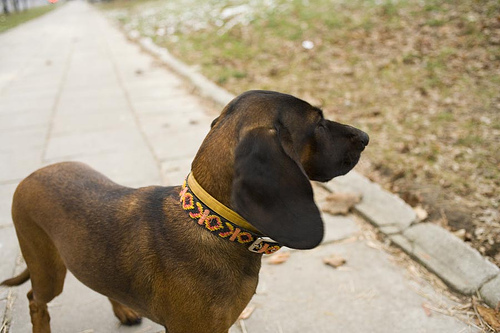}
%\caption{fig2}
\end{minipage}
}%
\hspace{.23in}
% \quad %这个回车键很重要 \quad也可以
\subfigure{
\begin{minipage}[t]{0.05\linewidth}
% \centering
\includegraphics[width=0.6in, height=0.6in]{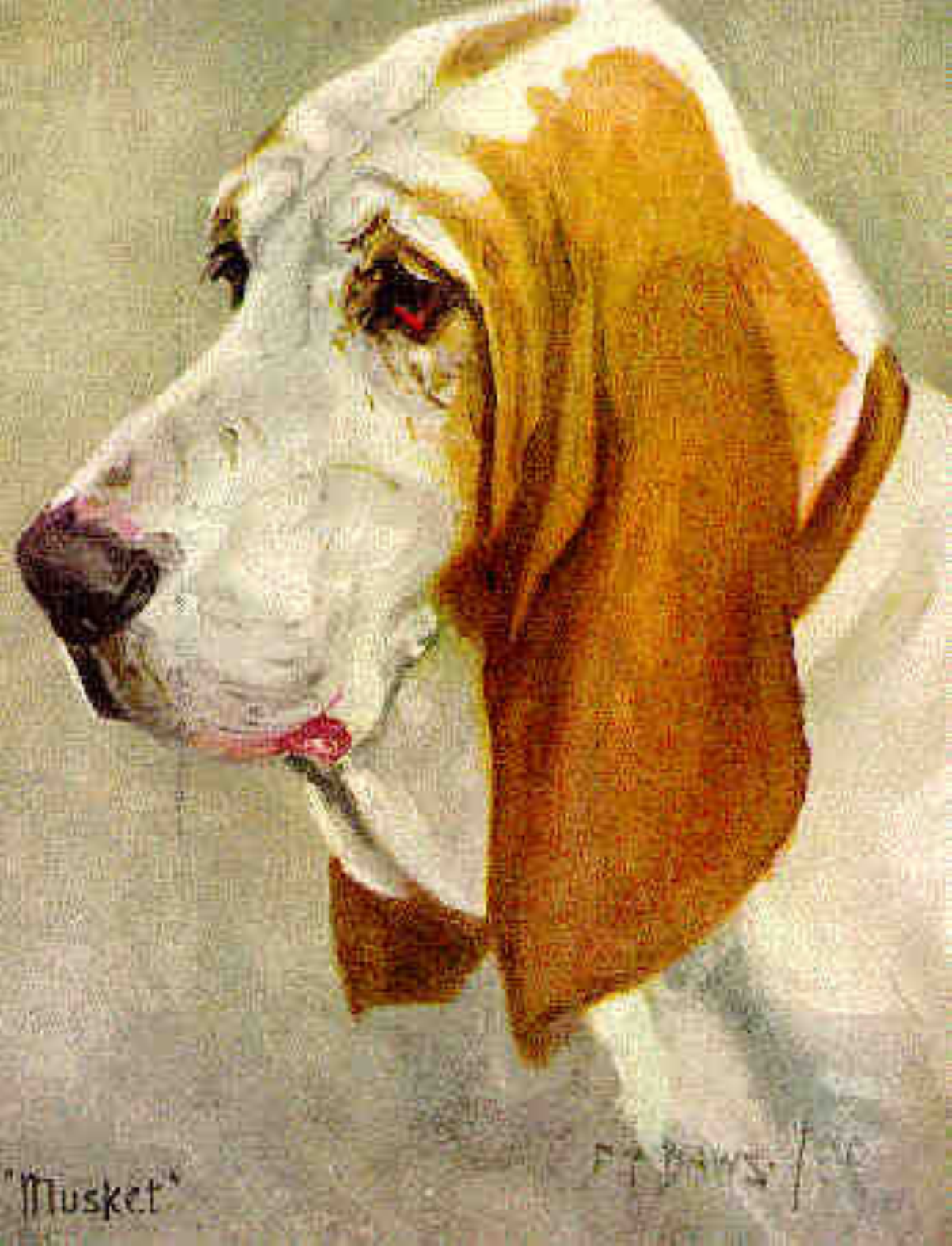}
%\caption{fig2}
\end{minipage}
}
\hspace{.20in}
\subfigure{
\begin{minipage}[t]{0.05\linewidth}
% \centering
\includegraphics[width=0.6in, height=0.6in]{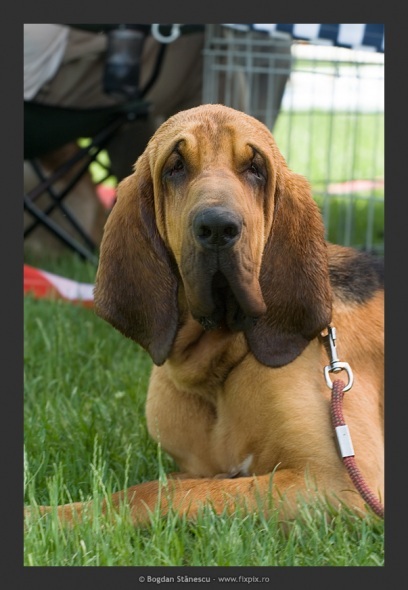}
%\caption{fig2}
\end{minipage}
}%
\hspace{.23in} 
\subfigure{
\begin{minipage}[t]{0.05\linewidth}
% \centering
\includegraphics[width=0.6in, height=0.6in]{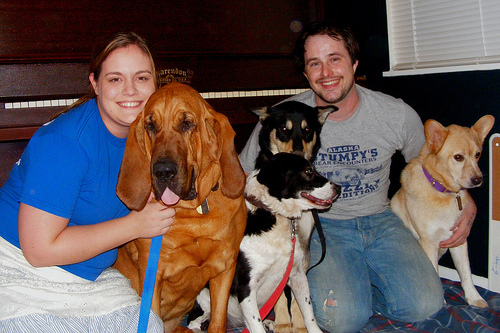}
%\caption{fig2}
\end{minipage}
}%
\hspace{.23in}
\subfigure{
\begin{minipage}[t]{0.05\linewidth}
% \centering
\includegraphics[width=0.6in, height=0.6in]{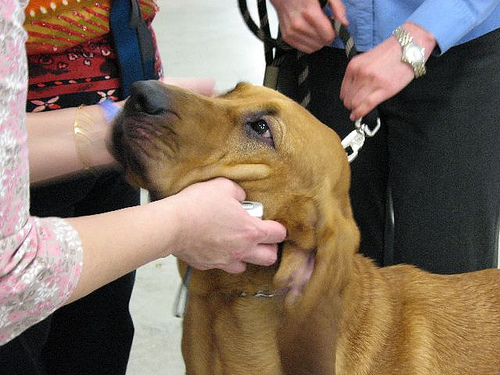}
%\caption{fig2}
\end{minipage}
}%
\hspace{.23in}
\subfigure{
\begin{minipage}[t]{0.05\linewidth}
% \centering
\includegraphics[width=0.6in, height=0.6in]{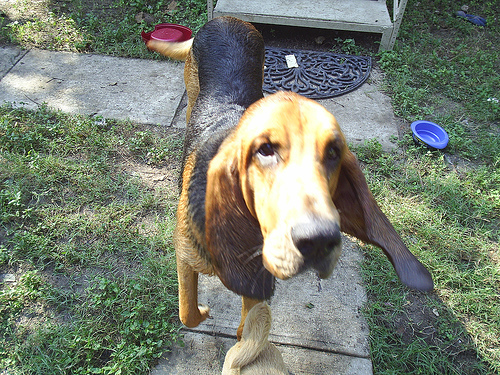}
%\caption{fig2}
\end{minipage}
}%
\hspace{.23in}
\subfigure{
\begin{minipage}[t]{0.05\linewidth}
% \centering
\includegraphics[width=0.6in, height=0.6in]{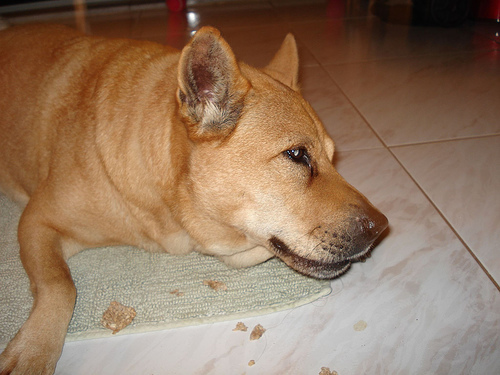}
%\caption{fig2}
\end{minipage}
}%
\hspace{.23in}
\subfigure{
\begin{minipage}[t]{0.05\linewidth}
% \centering
\includegraphics[width=0.6in, height=0.6in]{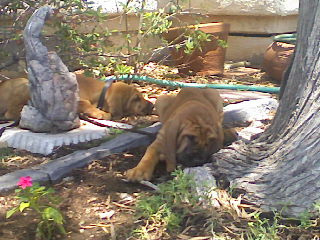}
%\caption{fig2}
\end{minipage}
}%
\hspace{.23in} 
\subfigure{
\begin{minipage}[t]{0.05\linewidth}
% \centering
\includegraphics[width=0.6in, height=0.6in]{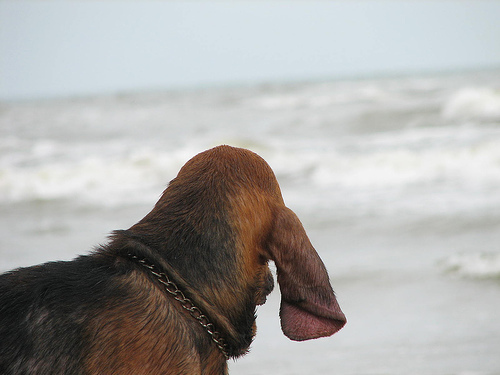}
%\caption{fig2}
\end{minipage}
}%

% \centering
\caption{The 4 classes randomly selected from ImageNet-100 validation set. For each class, we randomly show 10 figures. The top-down class numbers are n11939491, n01770081, n03179701, n02088466.}
\label{fig4}
\vskip -0.2in
\end{figure*}

\textbf{The influence of r.}
To balance the accuracy and cost of the model, we add a parameter r in the first FC. This is the same as SENet, but we use convolution operation to implement FC. From the table~\ref{tab5}, We can see that when r takes different values, it can balance accuracy and cost. Normally, when r is smaller, the parameters of the model increase and the accuracy of the model improves. 

\begin{table}[h]
\caption{The influence of r. The accuracy of the MobileNetV2 with different r on ImageNet-100.}\label{tab5}
\vskip 0.15in
\centering
\setlength{\tabcolsep}{2mm}{
\begin{tabular}{c|c|c|c|c|c|c}
\hline
\multicolumn{7}{c}{DN-B, batch size=64, lr=0.2, epoch=240} \\
\hline
r & g & \#p & Mult-Adds & Train & Val & Test \\
\hline
8& 1 & 3.72M & 301.02M & 91.20 & 85.30& 84.26 \\
16 & 1 & 3.03M & 300.34M & 91.62 & 85.63 & 84.08 \\
32 & 1 & 2.69M & 300.00M & 91.28& 85.34 & 83.36 \\
\hline
\end{tabular}}
\vskip -0.1in
\end{table}

\begin{table}[h]
\caption{The influence of g. The accuracy of the MobileNetV2 with different g on ImageNet-100.}\label{tab6}
\vskip 0.15in
\centering
\begin{threeparttable}
\setlength{\tabcolsep}{1.8mm}{
\begin{tabular}{c|c|c|c|c|c|c}
\hline
\multicolumn{7}{c}{DN-B, batch size=64, lr=0.2, epoch=240} \\
\hline
r & g & \#p & Mult-Adds & Train & Val & Test\\
\hline
16 & 1 & 3.03M & 300.34M & 91.62 & 85.63 & 84.08 \\
16 & 2 & 3.07M & 300.37M & 90.81& 84.08 & 82.42 \\
16 & 4 & 3.14M & 300.44M & 90.94 & 83.95 & 82.32 \\
16\tnote{1} & oup & 4.36M & 301.66M & 90.86 & 85.57 & 84.74 \\
\hline
\end{tabular}}
\begin{tablenotes}
       \footnotesize
       \item[1] The SC-Module in this experiment is implemented by two full connection operations.
     \end{tablenotes}
    \end{threeparttable}
% \vskip -0.1in
\end{table}

\textbf{The influence of g.}
Our second FC in SC-Module is implemented by group convolution operation. This is the same as WeightNet, but our g parameter represents that the number of channels in each group is g. From table~\ref{tab5}, we can see when g is the number of output channel, the model has the highest accuracy. When g is 1, the accuracy of the model with DN-B is still higher than that of the model with BN, and only a few parameters are added. 

\textbf{The influence of batch size.}
We also verified the robustness of DN-B under different batch sizes. When we reduce the batch size to 8 and train only 120 epochs, we find that the BN model can not train normally. However, the model using DN operation can maintain the accuracy for normal training. In addition, we also study the influence of batch size on the test accuracy of model using DN-B operation. We find that when the small batch size is used for model inference, the accuracy of model using DN-B decreases very little. For details of the results, see tabel~\ref{tab8}.

\begin{table}[h]
\caption{The influence of learning rate. The accuracy of the MobileNetV2 with different learning rate on ImageNet-100.}\label{tab7}
\vskip 0.15in
\centering
\setlength{\tabcolsep}{1mm}{
\begin{tabular}{c|c|c|c|c|c|c}
\hline
\multicolumn{7}{c}{DN-B, batch size=64, epoch=240} \\
\hline
BN/DN-B & Lr & \#p & Mult-Adds & Train & Val & Test \\
\hline
BN & 0.2 &2.35M & 299.62M & 86.48 & 83.61 & 82.50\\
BN & 0.5 &2.35M & 299.62M & 71.43 & 75.85 & 74.12 \\
DN(16, 1) & 0.2 & 3.03M & 300.34M & 91.62 & 85.63 & 84.08 \\
DN(16, 1) & 0.5 & 3.03M & 300.34M  & 88.69& 84.60 & 83.24 \\
\hline
\end{tabular}}
% \vskip -0.1in
\end{table}

\begin{figure*}[t]
\vskip 0.2in
\begin{center}
\centerline{\includegraphics[width=1\textwidth]{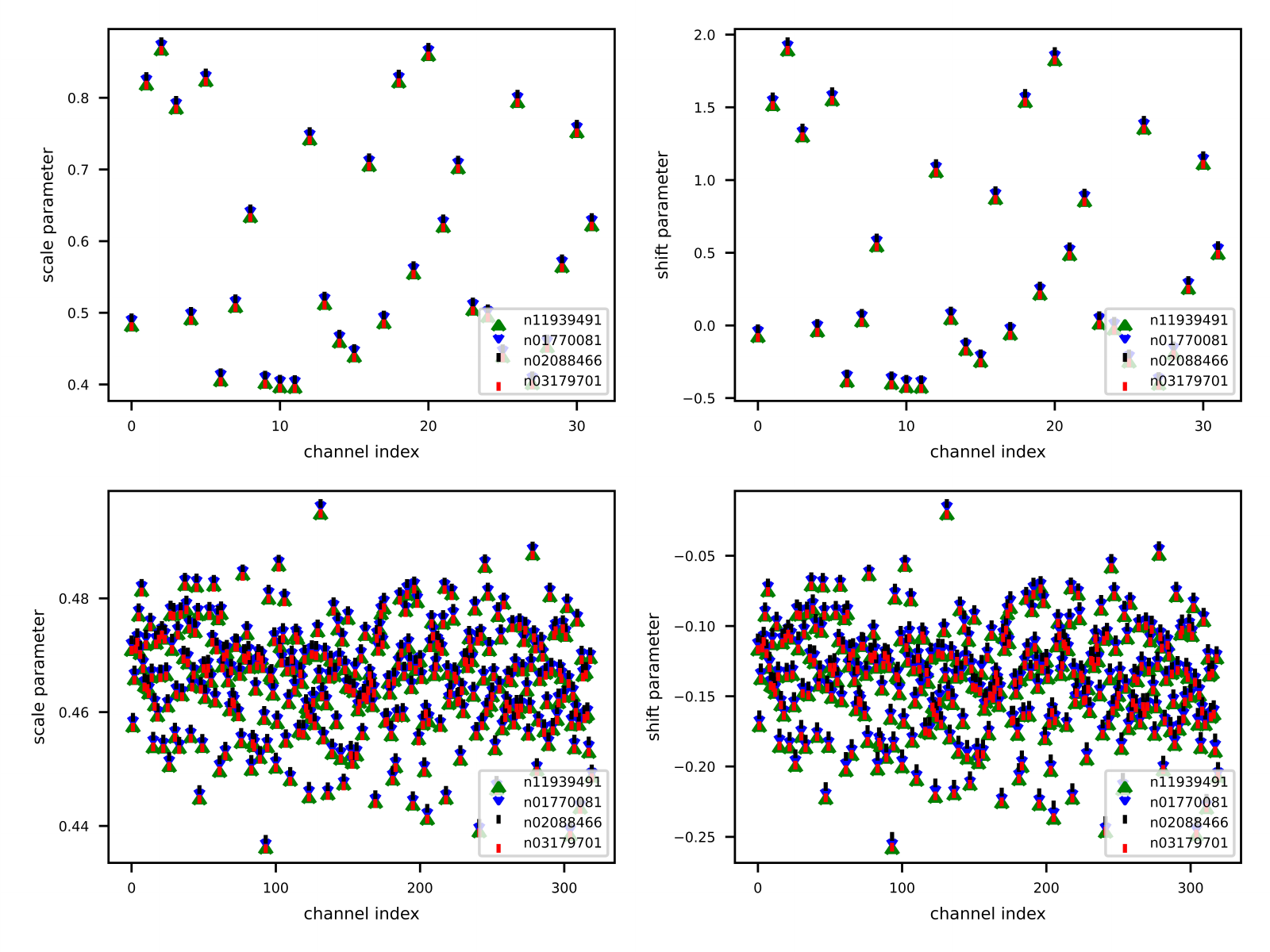}}
\caption{\textbf{The affine parameters} (best viewed in color). The top are the parameters of the DN-start operation in MobileNetV2 near the input. And the down are the parameters of the DN-end in MobileNetV2 near the output.}
\label{fig5}
\end{center}
\vskip -0.2in
\end{figure*}

\textbf{The influence of learning rate.}
As one of BN's advantages is that the network can use a relatively high learning rate. we carry out some experiments on a higher learning rate to verify the effectiveness of our method. When we increase the learning rate to 0.5, we find that the training accuracy of the model using BN decreases by 15+\%, and the test accuracy decreases by 8+\%. However, when the DN model also increases the learning rate to 0.5, the training accuracy of the model only decreases by 2+\%, and the test accuracy only decreases by 0.84\%. And the DN-B model with higher learning rate has better accuracy than the BN model with lower learning rate. 

\begin{table}[ht]
\caption{The test accuracy of the MobileNetV2 with DN-B on the 4 classes in figure \ref{fig4}.}\label{tab9}
\vskip 0.15in
\centering
\setlength{\tabcolsep}{1.5mm}{
\begin{tabular}{c|c|c|c|c}
\hline
\multicolumn{5}{c}{DN-B (16, oup) } \\
\hline
number & module & \#p & Mult-Adds & Test \\
\hline
n11939491 & DN(16,oup) & 4.36M & 301.66M & 96.00\\
n01770081 & DN(16,oup) & 4.36M & 301.66M & 90.00\\
n03179701 & DN(16,oup) & 4.36M & 301.66M & 74.00\\
n02088466 & DN(16,oup) & 4.36M & 301.66M & 68.00\\
\hline
\end{tabular}}
% \vskip -0.1in
\end{table}

\textbf{The visualization of Scale and Shift parameters.}
We randomly select 4 classes in ImageNet-100 and infer them based on MobileNetV2 with DN-B (16, 1) to get the parameters of affine operation. The 4 classes are shown in Figure \ref{fig4} and the test accuracy is shown in Table \ref{tab9}. The two DN operations we selected are at the beginning and the end of MobileNetV2, called DN-start and DN-end. From Figure \ref{fig5}, We find that the mean value of affine parameters between classes is almost the same, that is to say, DN can adaptively learn the function of BN in some layer of the network. Note that in DN-end operation, the number of channels is 1280. For visualization, we evenly display 320 channels at 4 intervals. 

% \begin{figure*}[ht]
% % \flushleft
% \vskip 0.2in
% \centering
% \subfigure{
% \begin{minipage}[t]{0.45\linewidth}
% % \centering
% \includegraphics[width=3in, height=2.9in]{2021ICML/start-scale.pdf}
% %\caption{fig2}
% \end{minipage}
% }%
% \hspace{0.23in} 
% \subfigure{
% \begin{minipage}[t]{0.45\linewidth}
% % \centering
% \includegraphics[width=3in, height=2.9in]{2021ICML/start-shift.pdf}
% %\caption{fig2}
% \end{minipage}
% }%
% \\
% % \hspace{0.23in} 
% \subfigure{
% \begin{minipage}[t]{0.45\linewidth}
% % \centering
% \includegraphics[width=3.1in, height=3in]{2021ICML/end-scale.pdf}
% %\caption{fig2}
% \end{minipage}
% }%
% \hspace{0.23in} 
% \subfigure{
% \begin{minipage}[t]{0.45\linewidth}
% % \centering
% \includegraphics[width=3.1in, height=3in]{2021ICML/end-shift.pdf}
% %\caption{fig2}
% \end{minipage}
% }%
% % \centering
% \caption{The 4 classes randomly selected from ImageNet-100 validation set. For each class, we randomly show 10 graphs. }
% \label{fig5}
% \vskip -0.2in
% \end{figure*}

\section{Discussion and Future Work}
In this work, we extend the static normalization to the dynamic normalization via SC-Module. We first extend normalization to generate parameters adaptively for mini-batch data (DN-C, Batch-shared and Channel-wise). Then, we extend DN-B to generate parameters adaptively for each channel of each sample (DN-B, Batch and Channel-wise). Our experiments show that DN-C model can't train normally, but DN-B model has very good robustness. And DN-B adds few parameters and gets better accuracy. In addition, compared with BN, DN-B has stable performance when using higher learning rate or smaller batch size. In this way, DN-B enhances the capability of BN and can be used as an alternative to BN in some cases. In the future, we will combine DN with Layer Norm \cite{ba2016layer}, Instance Norm \cite{ulyanov2016instance} and Group Norm \cite{wu2018group} to explore other possibilities of DN in RNN/LSTM or GAN models.

% \bibliography{example_paper}
% \bibliographystyle{icml2021}
% \clearpage

% \appendix
% \section{Do \emph{not} have an appendix here}

% \textbf{\emph{Do not put content after the references.}}
% %
% Put anything that you might normally include after the references in a separate
% supplementary file.

% We recommend that you build supplementary material in a separate document.
% If you must create one PDF and cut it up, please be careful to use a tool that
% doesn't alter the margins, and that doesn't aggressively rewrite the PDF file.
% pdftk usually works fine. 

% \textbf{Please do not use Apple's preview to cut off supplementary material.} In
% previous years it has altered margins, and created headaches at the camera-ready
% stage. 
%%%%%%%%%%%%%%%%%%%%%%%%%%%%%%%%%%%%%%%%%%%%%%%%%%%%%%%%%%%%%%%%%%%%%%%%%%%%%%%
%%%%%%%%%%%%%%%%%%%%%%%%%%%%%%%%%%%%%%%%%%%%%%%%%%%%%%%%%%%%%%%%%%%%%%%%%%%%%%%

\end{document}